\documentclass[numsec,webpdf,modern,large,namedate]{oup-authoring-template}

\usepackage{graphicx}
\graphicspath{{Fig/}}
\usepackage{dsfont}
\usepackage{hhline}
\usepackage{booktabs}
\setcitestyle{numbers}
\setcitestyle{square}

\setcitestyle{authoryear,round}

\usepackage[mathlines, switch]{lineno}

\theoremstyle{thmstyleone}%
\theoremstyle{thmstyletwo}%
\theoremstyle{thmstylethree}%

\begin{document}

\journaltitle{Bioinformatics}
\DOI{DOI HERE}
\copyrightyear{2026}
\pubyear{2014}
\access{Advance Access Publication Date: }
\appnotes{Paper}

\firstpage{1}

\subtitle{Subject Section}

\title[Adversarial Domain Adaptation Enables Knowledge Transfer Across Heterogeneous RNA-Seq Datasets]{Adversarial Domain Adaptation Enables Knowledge Transfer Across Heterogeneous RNA-Seq Datasets}

\author[1,2,$\ast$]{Kevin Dradjat}
\author[1]{Massinissa Hamidi}
\author[1]{Blaise Hanczar}

\authormark{Dradjat et al.}

\address[1]{\orgdiv{IBISC Laboratory}, \orgname{University Paris-Saclay (Univ. Evry)}, \orgaddress{\country{France}}}
\address[2]{\orgname{ADLIN}, \orgaddress{\postcode{Paris}, \country{France}}}

\corresp[$\ast$]{Corresponding author. \href{email:email-id.com}{kevin.dradjat@univ-evry.fr}}

\received{Date}{0}{Year}
\revised{Date}{0}{Year}
\accepted{Date}{0}{Year}


\abstract{
\textbf{Motivation:} Accurate phenotype prediction from RNA sequencing (RNA-seq) data is essential for diagnosis, biomarker discovery, and personalized medicine. Deep learning models have demonstrated strong potential to outperform classical machine learning approaches, but their performance relies on large, well-annotated datasets. In transcriptomics, such datasets are frequently limited, leading to over-fitting and poor generalization. Knowledge transfer from larger, more general datasets can alleviate this issue. However, transferring information across RNA-seq datasets remains challenging due to heterogeneous preprocessing pipelines and differences in target phenotypes.\\
\textbf{Results:} In this study, we propose a deep learning-based domain adaptation framework that enables effective knowledge transfer from a large general dataset to a smaller one for cancer type classification. The method learns a domain-invariant latent space by jointly optimizing classification and domain alignment objectives. To ensure stable training and robustness in data-scarce scenarios, the framework is trained with an adversarial approach with appropriate regularization. Both supervised and unsupervised approach variants are explored, leveraging labeled or unlabeled target samples. 
The framework is evaluated on three large-scale transcriptomic datasets (TCGA, ARCHS4, GTEx) to assess its ability to transfer knowledge across cohorts. Experimental results demonstrate consistent improvements in cancer and tissue type classification accuracy compared to non-adaptive baselines, particularly in low-data scenarios.
Overall, this work highlights domain adaptation as a powerful strategy for data-efficient knowledge transfer in transcriptomics, enabling robust phenotype prediction under constrained data conditions. \\
\textbf{Availability:} The code and results are available at \href{https://github.com/kdradjat/da_rnaseq}{github.com/kdradjat/da$\_$rnaseq} \\
\textbf{Contact:} \href{kevin.dradjat@univ-evry.fr}{kevin.dradjat@univ-evry.fr}\\
\textbf{Supplementary information:} Supplementary data are available at \textit{Bioinformatics}
online.}



\maketitle

\section{Introduction}

Accurate phenotype prediction from RNA sequencing (RNA-seq) data is crucial for advancing disease diagnosis, biomarker discovery, and personalized medicine. Deep learning models have recently shown strong potential for this task by capturing complex, non-linear dependencies  among genes that classical approaches cannot easily model \citep{DLomics}. However, the effectiveness of these models is highly dependent on the availability of large, well-annotated datasets. In practice, transcriptomic datasets are often limited in size, heterogeneous, and collected under diverse experimental or biological conditions, leading to over-fitting and limited generalization. These limitations motivate the use of transfer learning as a strategy to enhance the generalization of deep learning models and overcome data scarcity.

\subsection{Related Work}

\subsubsection{Transfer Learning for RNA-Seq data}

Transfer learning (TL) has been widely explored in computational biology to mitigate the impact of limited labeled data \citep{TL_survey}. TL typically involves pre-training deep learning models on a large source dataset and fine-tuning them on smaller target cohorts \citep{hanczar_assessment_2022}. Self-supervised learning approach follows the same paradigm but specifically uses unlabeled data and a pretext task to learn representations during the pre-training step \citep{ssrl_ged}. While these strategies can improve predictive performance, they implicitly assume that the pre-training and fine-tuning data follow similar distributions. In reality, RNA-seq data collected across studies often exhibit substantial distributional shifts. 
These shifts arise from technical variation, called batch effects, or unwanted biological variability due to sex, age, or other factors, acting as noise that reduces the efficiency of transfer across datasets.

\subsubsection{Domain Adaptation and Batch Effect Correction}

Domain adaptation (DA) provides a principled framework to overcome these limitations by explicitly accounting for distributional differences between datasets \citep{DA_review}. Rather than assuming alignment between source and target datasets, DA methods aim to learn a shared, domain-invariant representation in which samples from different datasets follow similar distributions. DA has achieved notable success in fields such as computer vision and natural language processing through adversarial and contrastive learning techniques \citep{DA_CV, DA_NLP}. However, DA remains underexplored in bulk RNA-seq analysis. Most existing applications focus on single-cell data \citep{sota_sc_1, sota_sc_2}, while only a limited number of studies consider bulk transcriptomics \citep{obsda}, often under strong distributional assumptions that restrict their applicability to small heterogeneous cohorts.

An alternative strategy involves statistical batch effect correction (BEC) methods, such as ComBat \citep{combat_ref} and limma \citep{limma_ref}, which aim to remove technical variation while preserving biological signals. Although widely used, these methods typically handle linear effects and may fail to capture complex non-linear shifts, especially when technical and biological variability interact. Adversarial DA provides a promising avenue to explore, in terms of the flexibility it offers to align source and target, potentially offering cross-dataset generalization for bulk RNA-seq tasks.

\subsection{Proposed Approach}
In this study, we develop a deep learning–based domain adaptation framework for phenotype prediction across heterogeneous RNA-seq datasets. The proposed framework allows to learn a domain-invariant latent space by jointly optimizing classification and alignment objectives, integrating adversarial learning with regularization strategies to ensure robust performance under data-scarce conditions. To further improve stability, we evaluate a Wasserstein-based discriminator for smoother domain alignment. We further develop supervised and unsupervised variants to handle varying levels of label availability in the target dataset.

We assess the proposed framework using three large-scale transcriptomics datasets (TCGA, ARCHS4, GTEx), each representing diverse experimental and biological conditions. Results show that our method effectively improves cancer and tissue type classification accuracy and enhances generalization across domains. Collectively, this work demonstrates that domain adaptation is a powerful and data-efficient strategy for knowledge transfer and phenotype prediction in transcriptomics.

\section{Material and Methods}
\subsection{Domain Adaptation}
Domain adaptation is a machine learning paradigm designed to transfer knowledge from a well-characterized source domain to a target domain with a different data distribution. The central objective of domain adaptation is to learn a representation that minimizes the discrepancy between source and target distributions, thereby enabling models trained on the source data to generalize effectively to the target data.

Formally, domain adaptation addresses the problem of transferring predictive models from a source domain $\mathcal{D}_{S}=\{(x_{i}^{S}, y_{i}^{S})\}_{i=1}^{N_{S}}$ with distribution $\mathcal{P}_{S}(x, y)$ to a target domain $\mathcal{D}_{T}\{(x_{i}^{T}, y_{i}^{T})\}_{i=1}^{N_{T}}$ with distribution $\mathcal{P}_{T}(x, y)$, where $\mathcal{P}_{S}(x) \neq \mathcal{P}_{T}(x)$ and $x_{i}$ correspond to data, $y_{i}$ to the labels and $N_{S}$ and $N_{T}$ respectively the number of source and target examples. The objective is to minimize the domain discrepancy while preserving discriminative features relevant to the task.

\begin{figure}[!h]
    \centering
    \includegraphics[scale=0.14]{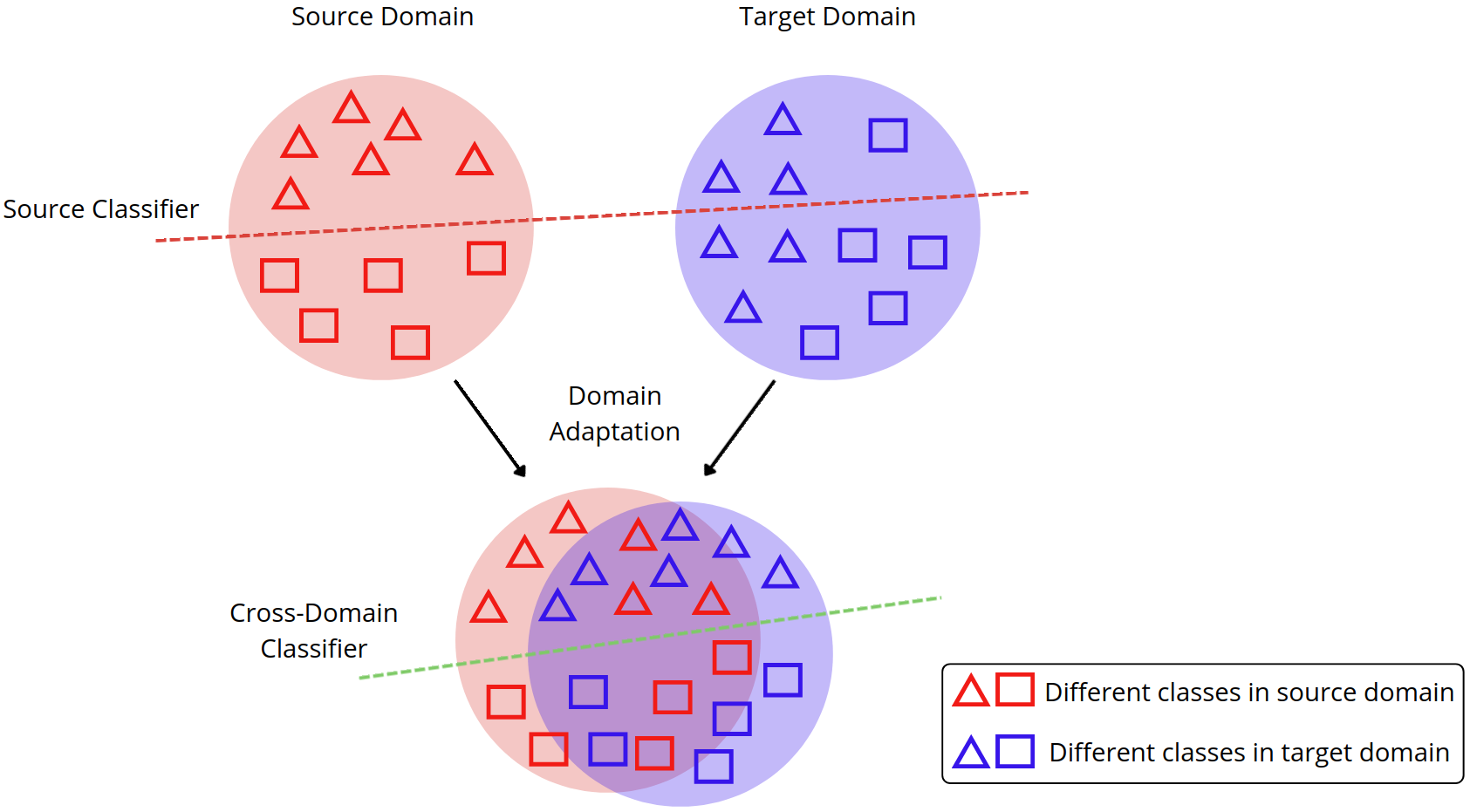}
    \caption{Conceptual illustration of domain adaptation process.}
    \label{DA_scheme}
\end{figure}

Depending on the availability of labels in the target domain, we consider two types of domain adaptation: Supervised DA, where a small set of labeled target samples is available to guide the alignment and unsupervised DA where target labels are completely unavailable.


In a classification context, DA aims to learn a classifier $C$ that achieves good performance on the target domain whose data distribution differs from the source domain. In this work, we focus on optimizing the target-performance of $C$, using source data for supervision signal and both domains for adaptation signal.
 
Classical approaches include moment-matching methods such as maximum mean discrepancy (MMD) minimization \citep{mmd_ref}, adversarial alignment using domain discriminators as in Domain-Adversarial Neural Networks (DANN) \citep{dann_ref}, and optimal transport–based techniques \citep{optimal_transport_ref}.
These approaches have become state-of-the-art in computer vision and text analysis by minimizing the divergence between source and target distributions in the latent space. Despite their success, the application of such methods to bulk RNA-seq data remains limited, where both technical batch effects and biological domain shifts (e.g., cancer vs. healthy or post-mortem tissues) create additional complexity \citep{bulk_challenges}.

\subsection{Architecture of the framework}
In this work, following recent advances in adversarial domain adaptation \citep{dann_ref, DA_wgan}, we adopt a deep learning framework that learns domain-invariant representations through joint optimization of classification and domain alignment objectives.

\begin{figure*}[!h]
    \centering
    \includegraphics[scale=0.71]{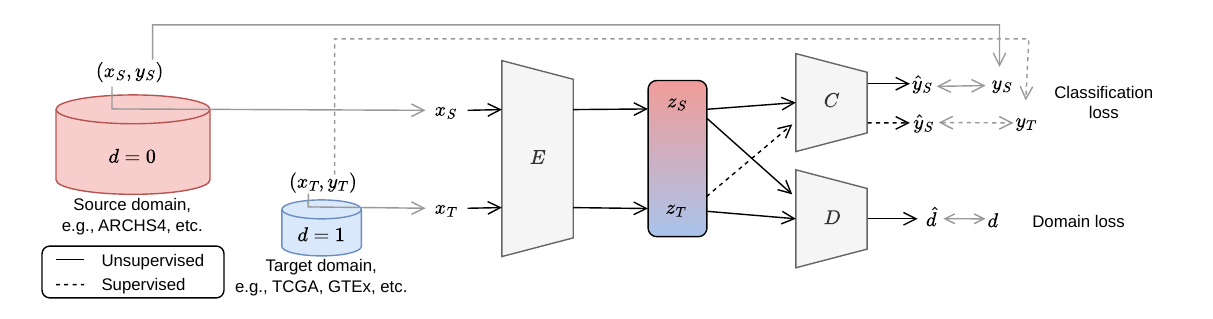}
    \caption{Composition of the domain adaptation pipeline. $E$, $C$ and $D$ denote, respectively, the encoder, the phenotype classifier and the domain discriminator. The unsupervised variant is defined by only the solid arrows, whereas we add the dotted arrows for the supervised variant. The discriminator loss can either be the Wasserstein distance or cross-entropy.}
    \label{archi}
\end{figure*}

The architecture, illustrated in Figure \ref{archi}, is composed of three main components. The encoder $E$ projects input samples $x$ into a low-dimensional latent space to produce representations $z$. The classifier $C$ predicts labels from these representations, while the discriminator $D$ aligns the distributions of source and target representations to encourage domain-invariant features.


\subsection{Domain adaptation objective}
We consider two settings depending on the availability of target labels. In the unsupervised case, only source labels are available: the encoder and classifier are trained using cross-entropy on the source data, while the discriminator is trained to distinguish source from target, and the encoder is optimized adversarially to fool it. In the supervised case, labels are available for both source and target samples. Here, the classifier is trained jointly on both domains, which reinforces alignment by ensuring class consistency across datasets, while the discriminator continues to provide adversarial guidance for domain invariance.


\textbf{Loss functions.} The overall training objective, balancing classification and domain alignment, is the following:
\begin{equation}
    \min_{E,C} \max_{D} \mathcal{L}_{cls}(E, C) + \lambda \mathcal{L}_{dom}(E, D),
\end{equation}
where $\lambda$ is a hyper-parameter that controls the strength of domain alignment. We choose the cross-entropy loss for $\mathcal{L}_{cls}$ and we define two alternatives for $\mathcal{L}_{dom}$: 
\begin{itemize}
    \item Cross-entropy domain loss (DANN):
    \begin{equation}
        \mathcal{L}_{dom} = -[d\log D(z) + (1-d)\log(1 - D(z))],
    \end{equation}
    where $z$ corresponds to the input representations and $d\in \{0,1\}$ indicates domain label.
    \item Wasserstein domain loss (Wasserstein-based) \citep{wgan_ref}:
    \begin{align} 
        \mathcal{L}_{dom} &= \mathbb{E}_{z\sim P_{S}}[D(z)] - \mathbb{E}_{z\sim P_{T}}[D(z)] \nonumber\\&+ \sigma\mathbb{E}_{\hat{z}}[(\Vert\nabla_{\hat{z}}D(\hat{z})\Vert_{2} - 1)^{2}], \label{adv_loss}
    \end{align}
    with gradient penalty applied to enforce the Lipschitz constraint, where $\hat{z}$ represents interpolated feature between source and target representations and $\sigma$ a hyper-parameter usually fixed to $10$. 
\end{itemize}

We therefore define four variants of the model depending on the domain discriminator loss used and whether we consider the target dataset labelled or not.

\begin{figure*}[!h]
    \centering
    \includegraphics[scale=0.34]{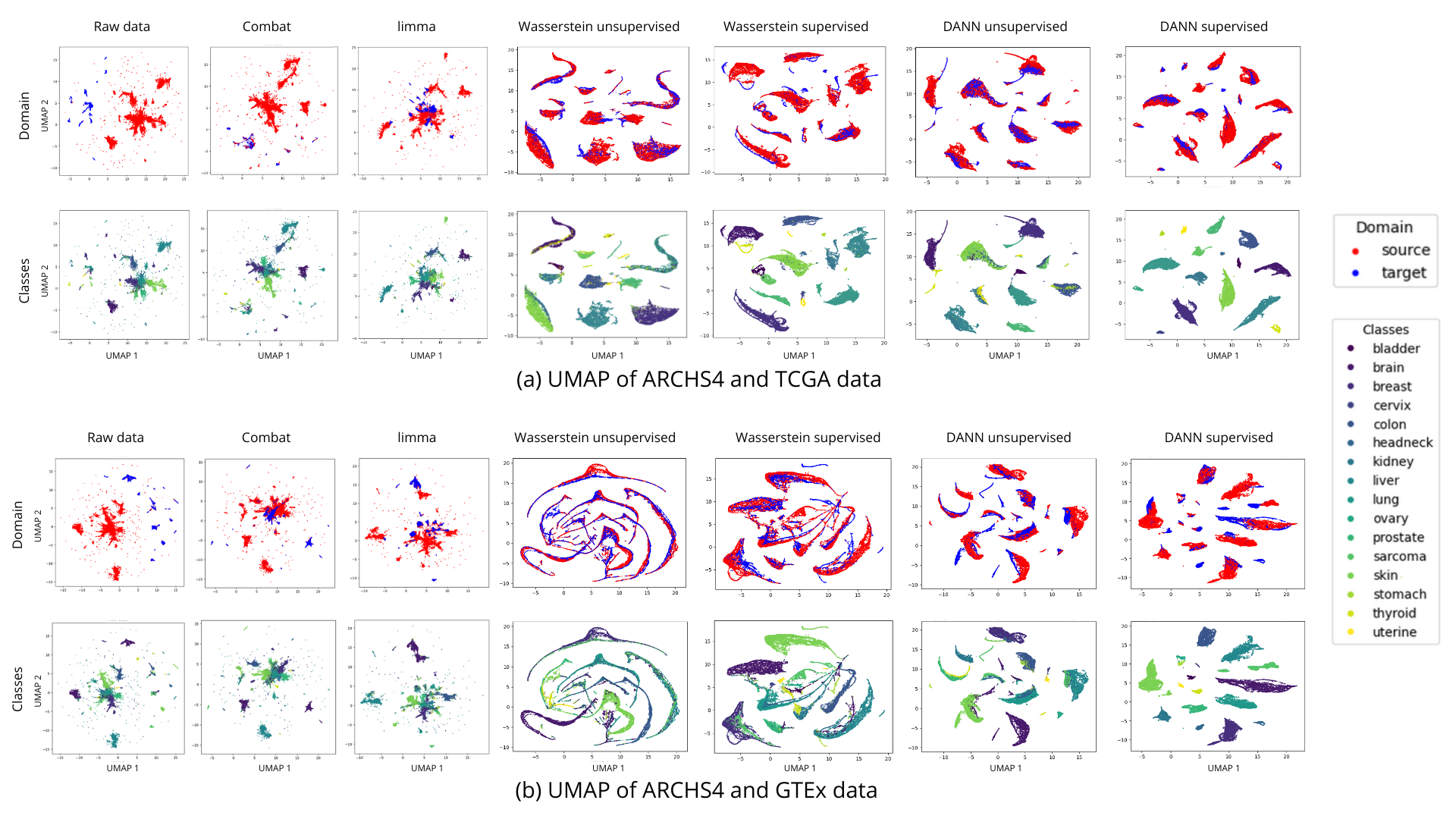}
    \caption{UMAP visualisations after domain adaptation for two study cases: (a) TCGA-target and (b) GTEx-target. 
             Top row shows embeddings coloured by domain and bottom row shows embeddings coloured by class. 
             (Top) TCGA dataset, (Bottom) GTEx dataset.}
    \label{umap_combined}
\end{figure*}

\section{Experiments and Results}
\subsection{Datasets}
We selected three RNA-Seq datasets, described in detail in the Supplementary Materials (Tables S1-S3), with the preprocessing method (Section A), to conduct our experiments: 
\begin{itemize}
    \item The Cancer Genome Atlas (TCGA): This is a pan-cancer RNA-Seq dataset that contains 9349 samples and 56902 input genes from 19 different types of cancer. 
    \item All RNA-Seq and ChIP-Seq Sample and Signature Search (ARCHS4) \citep{archs4}: This is a pan-tissue dataset originating from experiences from SRA (Sequence Read Archive) and GEO (Gene Expression Omnibus). 
    It contains 53282 samples from 19 different tissue types. 
    \item Genotype-Tissue Expression (GTEx) \citep{gtex}: This is a large-scale resource that profiles gene expression across multiple human tissues. It contains RNA sequencing data from postmortem tissue samples collected from hundreds of donors, covering 19 tissue types with 12962 samples. 
\end{itemize}

\subsection{Evaluation scenarios}
The goal is to evaluate the effectiveness of the designed domain adaptation methods with the TCGA, ARCHS4 and GTEx datasets. Although we defined supervised and unsupervised methods, all the datasets are labeled. We divided each dataset into a training, a validation and a test set while conserving initial class proportions. 
We consider two evaluation scenarios, treating one dataset as the source and another as the target in each case. In the first scenario, ARCHS4 serves as the source domain and TCGA as the target, with training sets of 45,763 and 7,291 samples, respectively, and test sets of 1,020 and 1,029 samples; 10\% of each training set was reserved for validation. In the second scenario, ARCHS4 is again the source while GTEx is the target, with training sets of 45,763 and 11,930 samples, test sets of 1,020 and 1,032 samples, and 10\% of the training sets used for validation.

\textbf{Model architecture and training.}
We selected the architecture by applying adaptation with the source and target training sets and taking the one with the best accuracy on the target validation set. The hyperparameters tested are listed in the Supplementary (Table S4). The encoder $E$ consists of four fully connected layers with 256 units each, using ReLU activations and batch normalization. We use a classifier $C$ with layer dimensions 128-64-$c$, where $c$ denotes the number of classes, and employs ReLU activations with batch normalization. The discriminator $D$ has a 256-128-64 architecture with LeakyReLU activations and a scalar output. For the domain adaptation part, we train the model using training sets from the source and target during 200 epochs with a batch size of 64 and a $\lambda$ hyperparameter fixed to 1.

\textbf{Benchmarking.} We apply each adaptation method to the different study cases and evaluate their performance against several baseline models. The first baseline is a multilayer perceptron (MLP) trained exclusively on target domain samples, referred to as "target-only". The second baseline corresponds to the proposed supervised approach without the adaptation component, referred to as "supervised no adaptation". Additionally, we compare our method with statistical batch effect correction (BEC) methods, including Combat and limma, that are commonly used to integrate datasets originating from multiple sources.
However, it is important to note that the datasets considered in this work differ not only due to batch effects but also as a consequence of distinct underlying biological contexts. 
\\

To comprehensively evaluate the proposed method, we considered three complementary scenarios. 
First, we analyze the structure of the learned latent space to examine whether domain adaptation effectively removes domain variability while preserving biologically meaningful phenotype structure. Second, we evaluate performance in a low target-data regime, reflecting clinical and rare-disease scenarios where only a limited number of target samples are available. Third, we assess robustness in a low source-data regime combined with a low target-data regime, where access to large reference datasets is restricted.

\subsubsection{Embeddings alignment} \label{umap_alignment}

This section investigates the structure of the latent representations learned by the model. For each case study, described above, we train the model with the entire source and target training sets. Then, we use UMAP for dimensionality reduction to visualize embeddings from both source and target domains. This experiment permits to highlight which method successfully removes domain-specific variation while preserving biologically relevant clustering according to cancer phenotypes. 

The results are presented in Figure \ref{umap_combined}a for the TCGA case and Figure \ref{umap_combined}b for the GTEx case. For the TCGA case, we distinguish the two different domains when observing raw data, indicating a pronounced distributional shift in the original data. After applying Combat and limma methods, partial mixing of domain is observed but clusters remain domain-dependant and class separation is blurred. The proposed domain adaptation models achieve stronger alignment than the baselines. In the unsupervised setting, source and target samples occupy overlapping regions of the latent space, indicating effective distributional alignment. The supervised variants further improve this alignment by simultaneously overlapping source and target embeddings and clustering samples by cancer class. Among the proposed methods, the supervised Wasserstein-based and supervised DANN models generate the most invariant and well-separated class clusters.

For the GTEx case, we observe the same trends in both the raw data distributions and the ComBat and limma batch-correction results. 
\clearpage
\begin{table*}
    \centering
    \begin{tabular}{cc c c cc c c c }
         \toprule
         \multirow{3}{*}{Method} & \multicolumn{4}{c}{non-adaptive} & \multicolumn{4}{c}{adaptive} \\
         \cmidrule(r){2-5}
         \cmidrule(r){6-9}
          & Target-only & Supervised  & Combat & limma & Wasserstein & Wasserstein & DANN & DANN \\
         & & no adaptation & & & unsupervised & supervised & unsupervised & supervised \\
         \hline
         TCGA & 98.76$\pm$0.1 & 91.91$\pm$0.1 & 97.88$\pm$0.086 & 97.56$\pm$0.05 & 73.13$\pm$0.72 & 97.18$\pm$0.17 & 63.95$\pm$1.82 & 96.94$\pm$0.25 \\
         \hline
         GTEx & 98.93$\pm$0.04 & 99.32$\pm$0.07 & 98.25$\pm$0.55 & 98.25$\pm$0.53 & 22.38$\pm$0.48 & 99.38$\pm$0.04 & 51.12$\pm$1.42 & 99.96$\pm$0.04 \\
         \bottomrule
    \end{tabular}
    \caption{Accuracy on the TCGA/GTEx target test set after each adaptation method when considering the entire source and target training sets.}
    \label{acc}
\end{table*}
\begin{figure*}
    \centering
    \includegraphics[scale=0.24]{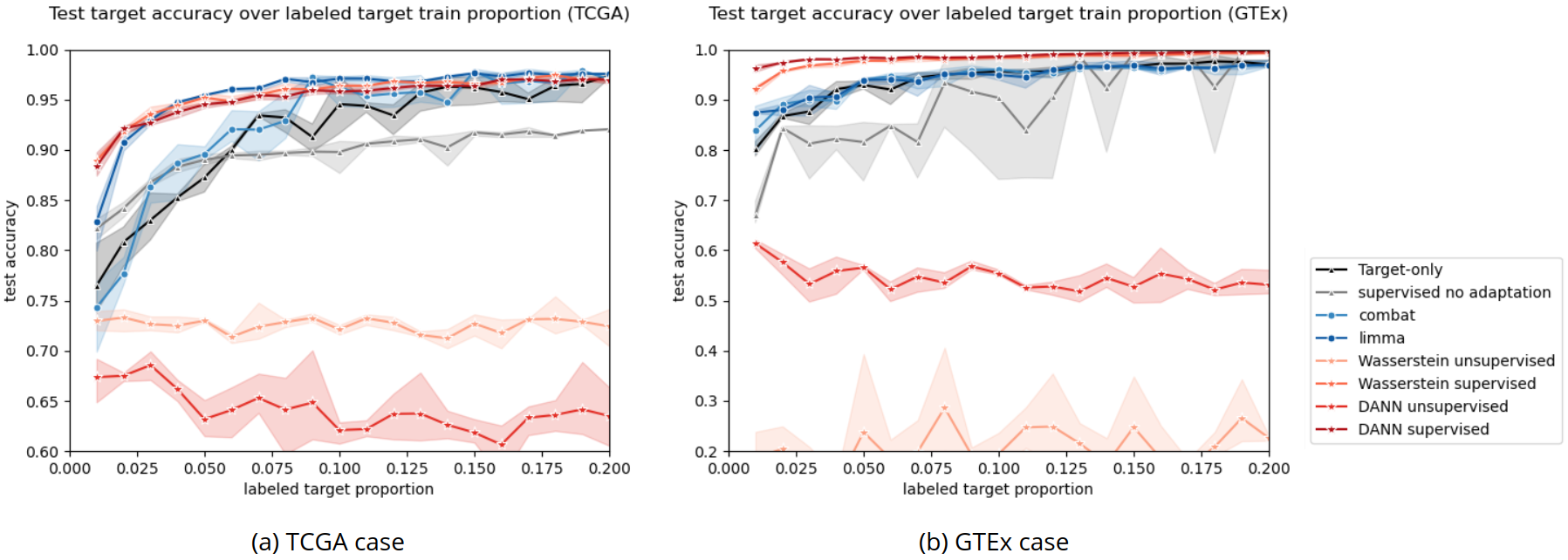}
    \caption{Performances of each adaptation methods by considering the TCGA dataset as target (a) and GTEx dataset as target (b).}
    \label{target_prop_acc}
\end{figure*}

The DANN approaches behave similarly: the unsupervised version aligns domains without aligning classes, while the supervised version produces domain-invariant embeddings with clear class separation.
Embeddings from the Wasserstein-based methods appear less structured than those from DANN, suggesting potential instability. They achieve good domain-level alignment, but class separation remains unclear for certain classes.
Overall, the results show that the proposed domain adaptation framework yields domain-invariant embeddings while preserving class structure, demonstrating its ability to capture underlying biological variation.


Then, since our main task is phenotype classification, we evaluate the accuracy of each considered method on the target test set. The results, presented in Table \ref{acc}, show that, excluding unsupervised methods (Wasserstein and DANN), all methods achieve high accuracy on this task. It therefore seems that domain adaptation methods bring no benefits in this case. However, the results were obtained by considering the entire target set, which represents enough data to achieve good classification accuracy with non-adaptive models. Besides, a high-data regime is not a common case and does not correspond to real-life cases.

\subsubsection{Low-target-data regime} \label{low_target}
The second experiment evaluates model performance when only a limited number of samples are available in the target domain. This setting mimics real-world scenarios such as rare cancer subtypes or small clinical cohorts, where baseline models usually do not achieve good performance. The goal is to assess the data efficiency of our approach and its ability to transfer predictive knowledge from large reference datasets to small, underrepresented target cohorts.

In this section, to simulate a low target data case, we apply each adaptation method using the entire source training set and with different proportion $p$ of target examples, from 0.01 to 0.2, with a step of 0.01. To estimate the variability of the approaches, we perform five iterations for each method and each target proportion by randomly initializing the model's weights. The same process was applied when evaluating the baselines.

The results are presented in Figure \ref{target_prop_acc}. 
For the TCGA case (Figure \ref{target_prop_acc}a), we can see that Combat batch effect correction achieves minor improvements over the target-only baseline, while limma clearly outperforms it when a considering few target proportion. The supervised no-adaptation baseline, trained jointly on source and labeled target without domain alignment, shows moderate improvements compared to the target-only baseline at low target proportion and ends up being inferior to it as the proportion increases. 

\paragraph{\textbf{Supervised Adaptation improves performances across all target proportions}}
Concerning the proposed methods, unsupervised and supervised approaches display different results. Both Wasserstein and DANN supervised adaptation methods significantly improve predictive performance across all target proportions, competing with limma batch effect correction. In contrast, unsupervised variants provide lower performance than baselines and supervised variants.
For the GTEx case (Figure \ref{target_prop_acc}b), we observe similar results. Indeed, Combat and limma achieve satisfying performance along with the target-only baseline, whereas the supervised no-adaptation method does not outperform these approaches. In the same way as the TCGA case, supervised adaptation methods achieve the best performances across all target proportions, while the unsupervised variants provide the lowest ones. 


Besides, we notice differences between TCGA and GTEx cases as the performances of supervised models seem higher for the GTEx case than for TCGA case. 

\begin{figure*}[t]
    \centering
    \includegraphics[scale=0.23]{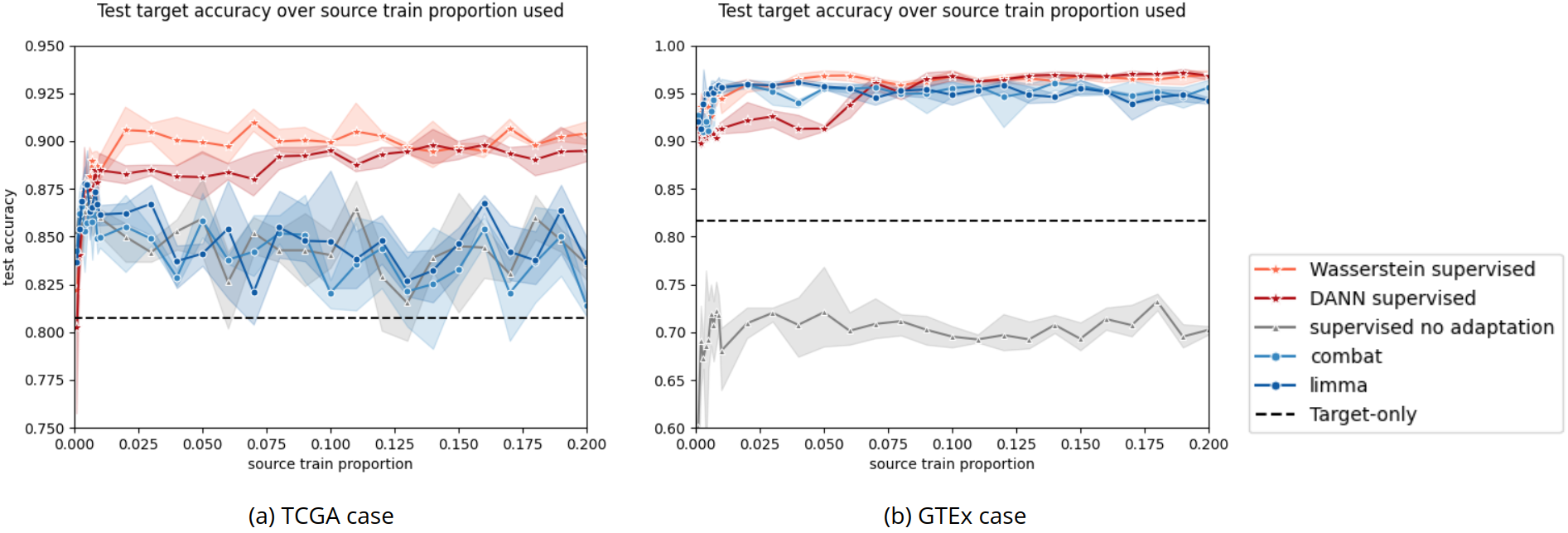}
    \caption{Test target accuracy for different proportion of source-domain training used for (a) TCGA and (b) GTEx targets.
Each model was trained with an increasing proportion of source samples while keeping labeled target proportion fixed to $0.01$ ($\sim$100 examples).}
    \label{source_prop_acc}
\end{figure*}

\subsubsection{Low-source-data regime} \label{low_source}
The third experiment examines the robustness of the framework when the amount of source data is restricted. Although large-scale public datasets are often used as source domains, access or size may be limited in certain cases, or subsampling may be necessary to ensure balanced comparisons. This experiment tests how well the model generalizes when trained with fewer source samples, providing insights into its scalability and resilience under data-limited conditions. In this section, we apply each adaptation method with a target proportion $p=0.01$. We vary the number of source examples used during the adaptation by increasing the proportion of source examples $q$ used from 0.001 to 0.01 with a step of 0.001, then from 0.01 to 0.02 with a step of 0.01. The results are presented in Figure \ref{source_prop_acc} for both target cases. Here, we focus on supervised approaches since these provide the best performance improvements in the low target regime (Section~\ref{low_target}). Results concerning unsupervised methods are available in the Supplementary (Figures S1, S2).
%

 Results reveal that our proposed approach is able to generalize well compared to classical batch correction methods and non-adaptive baselines. This observation is consistent across the two cases, TCGA and GTEx.
 For the TCGA case (Figure~\ref{source_prop_acc}a), the Wasserstein-supervised and DANN-supervised methods outperform the target-only baseline by constantly maintaining high accuracy across all source data proportions. Classical batch correction methods and non-adaptive baselines perform better than supervised adversarial methods under $q=0.05$. However, their accuracies decrease to plateau at a moderate accuracy under the supervised variants. 
For the GTEx target case (Figure \ref{source_prop_acc}b), we observe that the supervised no-adaptation baseline provides the lowest performance among all the methods, remaining below the target-only baseline. BEC methods provide the best gains even with a few source examples used. The supervised Wasserstein-based approach gives similar results across all source proportions, whereas the DANN-based method remains below BEC performances before reaching them at $q\approx0.7$. 


Intuitively, one would expect that using more source samples should necessarily improve knowledge transfer. However, for TCGA, the performance of BEC and non-adaptive approaches degrades as the source proportion grows, whereas our proposed method retains its generalization ability even as the source proportion increases.

\section{Discussion}
The results obtained with the three experiments on different study cases highlight the capacity of the proposed domain adaptation framework to effectively transfer knowledge across two different RNA-seq datasets.

Indeed, as seen in Section \ref{umap_alignment}, UMAP visualizations demonstrate that the latent representations learned through adversarial training successfully align source and target domains, whereas classical batch effect methods achieve partial harmonization. This shows that statistical approaches are not sufficient to correct the complex non-linear shifts observed between two gene expression datasets. In contrast, the proposed deep learning methods produce embeddings that are domain-invariant, which are also class-discriminative for supervised variants, showing that they maintain phenotype structure after alignment. However, unsupervised versions do not align phenotype representations, highlighting the importance of the classifier component in the adaptation process.

Then, we saw that the domain adaptation method provides satisfying predictive gains under target data-scarce conditions. Indeed, as seen in Section \ref{low_target}, when only a small proportion of labeled target samples is available, supervised domain adaptation improves cross-dataset accuracy compared to classical batch correction tools or non-adaptive baselines. This highlights the potential of combining adversarial alignment with minimal supervision to transfer predictive knowledge from a large reference dataset (ARCHS4) to smaller cohorts (TCGA or GTEx). This is particularly interesting in biomedical contexts, where obtaining large labeled datasets is often infeasible. 

Besides, we note a slightly superior predictive gain concerning the GTEx target case, which is accentuated in the low source data scenario. Indeed, the proposed methods provide predictive gain even under source data-scarce conditions, as seen in Section \ref{low_source}. Indeed, for the TCGA target case, both Wasserstein and DANN adaptation models maintain high accuracy, outperforming other baseline methods, demonstrating their ability to learn transferable representations from limited data in this case. In the GTEx target case, proposed adaptation models are equivalent to batch effect correction methods by providing higher accuracy gains, validating the fact that the adaptation seems easier when considering GTEx as the target. This can be explained by the smaller domain discrepancy between ARCHS4 and GTEx, as ARCHS4 aggregates large numbers of healthy tissue samples from public repositories, making its distribution more similar to GTEx, which also consists of healthy tissues. In contrast, TCGA is a cancer-related dataset, which increases the biological and technical divergence, making the alignment more challenging. In addition, this experiment demonstrates that, in the absence of explicit domain alignment, adding more source data does not consistently improve generalization. This finding supports the idea that minimizing domain discrepancy is essential for effective knowledge transfer rather than only increasing training volume. 

\section{Conclusion}
In this study, we introduced a deep learning–based domain adaptation framework for phenotype prediction across heterogeneous RNA-seq datasets. By jointly optimizing classification and domain alignment objectives, the method effectively mitigates both technical and biological sources of variability that typically hinder cross-study generalization. Comprehensive experiments demonstrate that the framework learns domain-invariant yet biologically meaningful representations, enabling accurate prediction even under limited data availability in both the source and target domains.

Overall, this work establishes domain adaptation as a powerful method to transfer knowledge from a source dataset to a smaller target dataset, capable of addressing both batch effects and biologically induced domain shifts. By aligning heterogeneous datasets within a shared latent space, the proposed framework improves phenotype prediction across studies and provides a scalable foundation for future applications in precision medicine, multi-cohort learning, and integrative omics analysis.




\section{Competing interests}
No competing interest is declared.

\section{Author contributions statement}
All authors conceived the presented idea and discussed the results and reviewed the manuscript. K.D redacted the manuscript and implemented the computations. 

\section{Funding}
This work has been supported by the Paris \^{I}le-de-France Région in the framework of DIM AI4IDF and by a collaboration with ADLIN Science.

\section{Acknowledgments}
The results shown here are in part based upon data generated by the TCGA Research Network: \href{https://www.cancer.gov/tcga}{https://www.cancer.gov/tcga}. We are grateful to the Institut Fran\c{c}ais de Bioinformatique for providing computing and storage resources.

\bibliographystyle{abbrvnat}
\bibliography{references}

\begin{thebibliography}{20}
\providecommand{\natexlab}[1]{#1}
\providecommand{\url}[1]{\texttt{#1}}
\expandafter\ifx\csname urlstyle\endcsname\relax
  \providecommand{\doi}[1]{doi: #1}\else
  \providecommand{\doi}{doi: \begingroup \urlstyle{rm}\Url}\fi

\bibitem[Aguet et~al.(2020)Aguet, Barbeira, Bonazzola, et~al.]{gtex}
F.~Aguet, A.~Barbeira, R.~Bonazzola, et~al.
\newblock The gtex consortium atlas of genetic regulatory effects across human tissues.
\newblock \emph{Science}, 369:\penalty0 1318--1330, 09 2020.

\bibitem[Boluki et~al.(2021)Boluki, Qian, and Dougherty]{obsda}
S.~Boluki, X.~Qian, and E.~R. Dougherty.
\newblock Optimal bayesian supervised domain adaptation for rna sequencing data.
\newblock \emph{Bioinformatics}, 37\penalty0 (19):\penalty0 3212--3219, 04 2021.
\newblock ISSN 1367-4803.

\bibitem[Courty et~al.(2017)Courty, Flamary, Tuia, and Rakotomamonjy]{optimal_transport_ref}
N.~Courty, R.~Flamary, D.~Tuia, and A.~Rakotomamonjy.
\newblock Optimal transport for domain adaptation.
\newblock \emph{IEEE Transactions on Pattern Analysis and Machine Intelligence}, 39\penalty0 (9):\penalty0 1853--1865, 2017.

\bibitem[Dradjat et~al.(2025)Dradjat, Hamidi, Bartet, and Hanczar]{ssrl_ged}
K.~Dradjat, M.~Hamidi, P.~Bartet, and B.~Hanczar.
\newblock Self-supervised representation learning on gene expression data.
\newblock \emph{Bioinformatics}, page btaf533, 09 2025.
\newblock ISSN 1367-4811.

\bibitem[Ganin et~al.(2016)Ganin, Ustinova, Ajakan, et~al.]{dann_ref}
Y.~Ganin, E.~Ustinova, H.~Ajakan, et~al.
\newblock Domain-adversarial training of neural networks.
\newblock \emph{J. Mach. Learn. Res.}, 17\penalty0 (1):\penalty0 2096–2030, Jan. 2016.
\newblock ISSN 1532-4435.

\bibitem[Gulrajani et~al.(2017)Gulrajani, Ahmed, Arjovsky, et~al.]{wgan_ref}
I.~Gulrajani, F.~Ahmed, M.~Arjovsky, et~al.
\newblock Improved training of wasserstein gans.
\newblock In \emph{Proceedings of the 31st International Conference on Neural Information Processing Systems}, NIPS'17, page 5769–5779, Red Hook, NY, USA, 2017. Curran Associates Inc.

\bibitem[Guo and Yu(2022)]{DA_NLP}
X.~Guo and H.~Yu.
\newblock On the domain adaptation and generalization of pretrained language models: A survey, 2022.
\newblock URL \url{https://arxiv.org/abs/2211.03154}.

\bibitem[Hanczar et~al.(2022)Hanczar, Bourgeais, and Zehraoui]{hanczar_assessment_2022}
B.~Hanczar, V.~Bourgeais, and F.~Zehraoui.
\newblock Assessment of deep learning and transfer learning for cancer prediction based on gene expression data.
\newblock \emph{BMC Bioinformatics}, 23\penalty0 (262), 2022.

\bibitem[Lachmann et~al.(2018)]{archs4}
A.~Lachmann et~al.
\newblock Massive mining of publicly available {RNA}-seq data from human and mouse.
\newblock \emph{Nature Communications}, 9:\penalty0 1366, 2018.

\bibitem[Liu et~al.(2022)Liu, Yoo, Xing, et~al.]{DA_review}
X.~Liu, C.~Yoo, F.~Xing, et~al.
\newblock Deep unsupervised domain adaptation: A review of recent advances and perspectives.
\newblock \emph{APSIPA Transactions on Signal and Information Processing}, 05 2022.

\bibitem[Pudova et~al.(2025)Pudova, Pavlov, Guvatova, et~al.]{bulk_challenges}
E.~A. Pudova, V.~S. Pavlov, Z.~G. Guvatova, et~al.
\newblock Machine learning models for cancer research: A narrative review of bulk rna-seq applications.
\newblock \emph{International Journal of Molecular Sciences}, 26\penalty0 (24), 2025.

\bibitem[Ritchie et~al.(2015)Ritchie, Phipson, Wu, et~al.]{limma_ref}
M.~E. Ritchie, B.~Phipson, D.~Wu, et~al.
\newblock limma powers differential expression analyses for rna-sequencing and microarray studies.
\newblock \emph{Nucleic Acids Research}, 43\penalty0 (7):\penalty0 e47--e47, 01 2015.

\bibitem[Shen et~al.(2018)Shen, Qu, Zhang, and Yu]{DA_wgan}
J.~Shen, Y.~Qu, W.~Zhang, and Y.~Yu.
\newblock Wasserstein distance guided representation learning for domain adaptation.
\newblock AAAI Press, 2018.

\bibitem[Sun and Qiu(2023)]{sota_sc_1}
Y.~Sun and P.~Qiu.
\newblock Domain adaptation for supervised integration of scrna-seq data.
\newblock \emph{Communications Biology}, 6\penalty0 (1):\penalty0 274, 2023.

\bibitem[Tran et~al.(2021)]{DLomics}
K.~Tran et~al.
\newblock Deep learning in cancer diagnosis, prognosis and treatment selection.
\newblock \emph{Genome Medicine}, 13\penalty0 (152), 2021.

\bibitem[Wang et~al.(2025)Wang, Wang, Zhao, et~al.]{TL_survey}
H.~Wang, J.~Wang, Z.~Zhao, et~al.
\newblock Understanding knowledge transferability for transfer learning: A survey, 2025.
\newblock URL \url{https://arxiv.org/abs/2507.03175}.

\bibitem[Wang and Deng(2018)]{DA_CV}
M.~Wang and W.~Deng.
\newblock Deep visual domain adaptation: A survey.
\newblock \emph{Neurocomput.}, 312\penalty0 (C):\penalty0 135–153, Oct. 2018.
\newblock ISSN 0925-2312.

\bibitem[Yan et~al.(2017)Yan, Ding, Li, et~al.]{mmd_ref}
H.~Yan, Y.~Ding, P.~Li, et~al.
\newblock Mind the class weight bias: Weighted maximum mean discrepancy for unsupervised domain adaptation.
\newblock In \emph{2017 IEEE Conference on Computer Vision and Pattern Recognition (CVPR)}, pages 945--954, 2017.

\bibitem[Zhang et~al.(2020)Zhang, Parmigiani, and Johnson]{combat_ref}
Y.~Zhang, G.~Parmigiani, and W.~E. Johnson.
\newblock Combat-seq: batch effect adjustment for rna-seq count data.
\newblock \emph{NAR Genomics and Bioinformatics}, 2\penalty0 (3):\penalty0 lqaa078, 09 2020.

\bibitem[Zhao et~al.(2025)Zhao, Song, Wei, et~al.]{sota_sc_2}
B.~Zhao, K.~Song, D.-Q. Wei, et~al.
\newblock sccobra allows contrastive cell embedding learning with domain adaptation for single cell data integration and harmonization.
\newblock \emph{Communications Biology}, 8, 02 2025.

\end{thebibliography}


\end{document}